\documentclass{article} % For LaTeX2e
\usepackage{iclr2015,times}
\usepackage{hyperref}
\usepackage{url}

\usepackage{amsmath}
\usepackage{amssymb}
\usepackage{amstext}

\usepackage[color=blue!40]{todonotes}

\usepackage{caption}

\usepackage{subcaption}

\usepackage{selectp}
\usepackage{sidecap}

\usepackage{amsfonts}
\usepackage{amsthm}
\usepackage{dsfont}

\DeclareMathOperator{\tr}{tr}

\title{Word Representations via \\Gaussian Embedding}

\author{
Luke Vilnis, Andrew McCallum\\
School of Computer Science \\
University of Massachusetts Amherst \\
Amherst, MA 01003 \\
\texttt{luke@cs.umass.edu, mccallum@cs.umass.edu} \\
}

\iclrfinalcopy % Uncomment for camera-ready version

\iclrconference % Uncomment if submitted as conference paper instead of workshop

\begin{document}

\maketitle

\begin{abstract}
Current work in lexical distributed representations maps each word to
a {\sl point} vector in low-dimensional space.  Mapping instead to a
{\sl density} provides many interesting advantages, including
better capturing uncertainty about a representation and its
relationships, expressing asymmetries more naturally than dot
product or cosine similarity, and enabling more expressive
parameterization of decision boundaries.  This paper advocates for
density-based distributed embeddings and presents a method for learning
representations in the space of Gaussian distributions. We compare 
performance on various word embedding benchmarks, investigate the 
ability of these embeddings to model entailment and other asymmetric 
relationships, and explore novel properties of the representation.
\end{abstract}

\section{Introduction}

In recent years there has been a surge of interest in learning compact
\emph{distributed representations} or \emph{embeddings} for many
machine learning tasks, including collaborative filtering 
\citep{Koren:2009:MFT:1608565.1608614}, image retrieval
\citep{weston2011wsabie}, relation extraction
\citep{riedel2013relation}, word semantics and language modeling
\citep{bengio2006neural,mnih2008scalable,mikolov2013distributed}, and many others.
In these approaches input objects (such as images, relations or words)
are mapped to dense vectors having lower-dimensionality than the
cardinality of the inputs, with the goal that the geometry of his
low-dimensional latent embedded space be smooth with respect to some
measure of similarity in the target domain.  That is, objects
associated with similar targets should be mapped to nearby points in
the embedded space.

While this approach has proven powerful, representing an object as a
single {\sl point} in space carries some important limitations.  An
embedded vector representing a point estimate does not naturally
express uncertainty about the target concepts
with which the input may be associated.  Point vectors are typically
compared by dot products, cosine-distance or Euclean distance, none of
which provide for asymmetric comparisons between objects (as is
necessary to represent inclusion or entailment).  Relationships
between points are normally measured by distances required to obey
the triangle inequality.

This paper advocates moving beyond vector {\sl point} representations
to \emph{potential functions} \citep{Aizerman67theoretical}, or continuous densities in latent
space.  In particular we explore Gaussian function embeddings (currently
with diagonal covariance), in which both means and variances are
learned from data.  Gaussians innately represent uncertainty, and provide a distance function per object.
KL-divergence between Gaussian distributions is straightforward to
calculate, naturally asymmetric, and has a geometric interpretation as an inclusion 
between families of ellipses.

There is a long line of previous work in mapping data cases to
probability distributions, perhaps the most famous being radial basis
functions (RBFs), used both in the kernel and neural network
literature.  We draw inspiration from this work to propose novel word
embedding algorithms that embed words directly as Gaussian
distributional potential functions in an infinite dimensional function
space. This allows us to map word types not only to vectors but to
soft regions in space, modeling uncertainty, inclusion, and entailment, as
well as providing a rich geometry of the latent space.

After discussing related work and presenting our algorithms below we
explore properties of our algorithms with multiple qualitative and
quantitative evaluation on several real and synthetic datasets.  We
show that concept containment and specificity matches common intuition on examples
concerning people, genres, foods, and others.  We compare our embeddings to Skip-Gram
on seven standard word similarity tasks, and evaluate the ability of our method to
learn unsupervised lexical entailment. We also demonstrate that our
training method also supports new styles of supervised training that
explicitly incorporate asymmetry into the objective.

\section{Related Work}

This paper builds on a long line of work on both distributed and
distributional semantic word vectors, including distributional
semantics, neural language models, count-based language models, and,
more broadly, the field of representation learning.

Related work in probabilistic matrix factorization
\citep{mnih2007probabilistic} embeds rows and columns as Gaussians, and some forms of this 
do provide each row and column with its own variance \citep{salakhutdinov2008bayesian}. Given the parallels
between embedding models and matrix factorization \citep{deerwester1990,riedel2013relation,levy2014neural}, this is relevant to our approach. However, these Bayesian methods apply Bayes' rule to observed data to infer the latent distributions, whereas our model works directly in the space of probability distributions and discriminatively trains them. This allows us to go beyond the Bayesian approach and use arbitrary (and even asymmetric) training criteria, and is more similar to methods that learn kernels \citep{lanckriet2004learning} or function-valued neural networks such as mixture density networks \citep{bishop1994mixture}.

Other work in multiplicative tensor
factorization for word embeddings \citep{kiros2014multiplicative} and
metric learning \citep{xing2002distance} learns some combinations of
representations, clusters, and a distance metric jointly;
however, it does not effectively
learn a distance function per item. Fitting Gaussian mixture models on embeddings 
has been done in order to apply Fisher kernels to entire documents \citep{clinchant2013aggregating,Clinchant:2013:TSB:2499178.2499180}. 
Preliminary concurrent work from
\citet{kiyoshiyo2014representing} describes a significantly
different model similar to Bayesian matrix factorization, using a probabilistic Gaussian graphical model to define a
distribution over pairs of words, and they lack quantitative
experiments or evaluation.  

In linguistic semantics, work on 
the \emph{distributional inclusion hypothesis}
\citep{geffet2005distributional}, uses traditional count-based vectors
to define regions in vector space \citep{erk2009representing} such that
subordinate concepts are included in these regions. In fact, one
strength of our proposed work is that we extend these intuitively appealing
ideas (as well as the ability to use a variety of asymmetric distances
between vectors) to the dense, low-dimensional distributed vectors
that are now gaining popularity.

\begin{SCfigure}
\centering
\caption{\label{fig-ellipses}Learned diagonal variances, as used in evaluation (Section \ref{evaluation}), for each word,
  with the first letter of each word indicating the position of its
  mean. We project onto generalized eigenvectors between the mixture means and variance of query word \emph{Bach}. Nearby words to \emph{Bach} are other composers e.g. \emph{Mozart}, which lead to similar pictures.}
\includegraphics[width=0.6\textwidth]{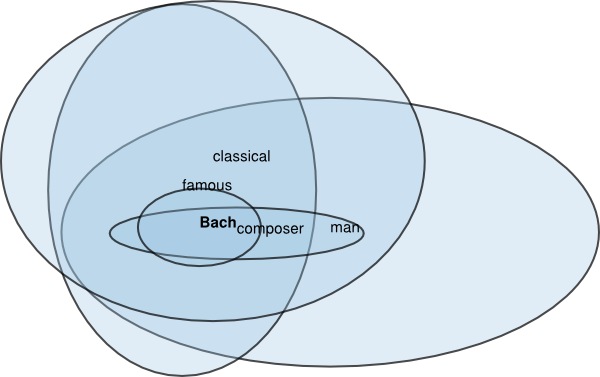}

\end{SCfigure}

\section{Background}
\label{background-section}

Our goal is to map every word type $w$ in some dictionary $\mathcal{D}$ and context word type $c$ in a dictionary $\mathcal{C}$ to a Gaussian distribution over a latent embedding space, such
that linguistic properties of the words are captured by properties of and relationships between the distributions. For precision, we call an element of the dictionary a \emph{word type}, and a particular observed token in some context a \emph{word token}. This is analogous to the \emph{class} vs. \emph{instance} distinction in object-oriented programming.

In unsupervised learning of word vectors, we observe a sequence of word tokens $\{t(w)_i\}$ for each type $w$, and their contexts (sets of nearby word tokens), $\{c(w)_i\}$. The goal is to map each word type $w$ and context word type $c$ to a vector, such that types that appear in similar contexts have similar vectors. When it is unambiguous, we also use the variables $w$ and $c$ to denote the vectors associated to that given word type or context word type.

An \emph{energy function} \citep{lecun2006tutorial} is a function $E_\theta(x,y)$ that scores pairs of inputs $x$ and outputs $y$, parametrized by $\theta$. The goal of \emph{energy-based learning} is to train the parameters of the energy function to score observed positive input-output pairs higher (or lower, depending on sign conventions) than negative pairs. This is accomplished by means of a \emph{loss function} $L$ which defines which pairs are positive and negative according to some supervision, and provides gradients on the parameters given the predictions of the energy function.

In prediction-based (energy-based) word embedding models, the parameters $\theta$ correspond to our learned word representations, and the $x$ and $y$ input-output pairs correspond to word tokens and their contexts. These contexts can be either positive (observed) or negative (often randomly sampled). In the word2vec Skip-Gram \citep{mikolov2013distributed} word embedding model, the energy function takes the form of a dot product between the vectors of an observed word and an observed context $w^\top c$. The loss function is a binary logistic regression classifier that treats the score of a word and its observed context as the score of a positive example, and the score of a word and a randomly sampled context as the score of a negative example.

Backpropagating \citep{RumelhartHintonWilliams1986} this loss to the word vectors trains them to be predictive of their contexts, achieving the desired effect (words in similar contexts have similar vectors). In recent work, word2vec has been shown to be equivalent to factoring certain types of weighted pointwise mutual information matrices \citep{levy2014neural}.

In our work, we use a slightly different loss function than Skip-Gram word2vec embeddings. Our energy functions take on a more limited range of values than do vector dot products, and their dynamic ranges depend in complex ways on the parameters. Therefore, we had difficulty using the word2vec loss that treats scores of positive and negative pairs as positive and negative examples to a binary classifier, since this relies on the ability to push up on the energy surface in an absolute, rather than relative, manner. To avoid the problem of absolute energies, we train with a ranking-based loss. We chose a max-margin ranking objective, similar to that used in Rank-SVM \citep{joachims2002optimizing} or Wsabie \citep{weston2011wsabie}, which pushes scores of positive pairs above negatives by a margin:
\begin{equation*}
L_m(w,c_p,c_n) =  \max(0, m - E(w, c_p) + E(w, c_n))
\end{equation*}
In this terminology, the contribution of our work is a pair of energy functions for training Gaussian distributions to represent word types.

\section{Warmup: Empirical Covariances}

Given a pre-trained set of word embeddings trained from contexts, there is a simple way to construct variances using the empirical variance of a word type's set of context vectors.

For a word $w$ with $N$ word vector sets $\{c(w)_i\}$ representing the words found in its contexts, and window size $W$, the empirical variance is
\begin{equation*}
\Sigma_w=\frac{1}{NW}\sum_i^N \sum_{j}^W (c(w)_{ij} - w) (c(w)_{ij} - w)^\top
\end{equation*}

This is an estimator for the covariance of a distribution assuming that the mean is fixed at $w$. In practice, it is also necessary to add a small \emph{ridge} term $\delta > 0$ to the diagonal of the matrix to regularize and avoid numerical problems when inverting.

However, in Section \ref{entailment-experiments} we note that the distributions learned by this empirical estimator do not possess properties that we would want from Gaussian distributional embeddings, such as unsupervised entailment represented as inclusion between ellipsoids. By discriminatively embedding our predictive vectors in the space of Gaussian distributions, we can improve this performance. Our models can learn certain forms of entailment during unsupervised training, as discussed in Section \ref{entailment-experiments} and exemplified in Figure \ref{fig-ellipses}.

\section{Energy-Based Learning of Gaussians}

As discussed in Section \ref{background-section}, our architecture learns Gaussian distributional embeddings to predict words in context given the current word, and ranks these over negatively sampled words. We present two energy functions to train these embeddings.

\subsection{Symmetric Similarity: Expected Likelihood or Probability Product Kernel}

While the dot product between two means of independent Gaussians is a perfectly valid measure of similarity (it is the expected dot product), it does not incorporate the covariances and would not enable us to gain any benefit from our probabilistic model.

The most logical next choice for a symmetric similarity function would be to take the inner product between the distributions themselves. Recall that for two (well-behaved) functions $f, g \in \mathbb{R}^n \to \mathbb{R}$, a standard choice of inner product is
\begin{equation*}
\int_{x \in \mathbb{R}^n} f(x) g(x) dx
\end{equation*}
i.e. the continuous version of $\sum_i f_i g_i=\langle f, g \rangle$ for discrete vectors $f$ and $g$.

This idea seems very natural, and indeed has appeared before -- the idea of mapping data cases $w$ into probability distributions (often over their contexts), and comparing them via integrals has a history under the name of the \emph{expected likelihood} or \emph{probability product kernel} \citep{jebara2004probability}.

For Gaussians, the inner product is defined as
\begin{equation*}
E(P_i,P_j)=\int_{x \in \mathbb{R}^n}\mathcal{N}(x;\mu_i,\Sigma_i)\mathcal{N}(x;\mu_j,\Sigma_j)dx = \mathcal{N}(0;\mu_i - \mu_j,\Sigma_i + \Sigma_j)
\end{equation*}
The proof of this identity follows from simple calculus. This is a consequence of the broader fact that the Gaussian is a \emph{stable distribution}, i.e. the convolution of two Gaussian random variables is another Gaussian.

Since we aim to discriminatively train the weights of the energy function, and it is always positive, we work not with this quantity directly, but with its logarithm. This has two motivations: firstly, we plan to use ranking loss, and ratios of densities and likelihoods are much more commonly worked with than differences -- differences in probabilities are less interpretable than an odds ratio. Secondly, it is easier numerically, as otherwise the quantities can get exponentially small and harder to deal with.

The logarithm of the energy (in $d$ dimensions) is
\begin{equation*}
\log \mathcal{N}(0;\mu_i - \mu_j,\Sigma_i + \Sigma_j) = -\frac{1}{2}\log \det(\Sigma_i + \Sigma_j) - \frac{1}{2}(\mu_i - \mu_j)^\top (\Sigma_i + \Sigma_j)^{-1} (\mu_i - \mu_j) - \frac{d}{2}\log(2\pi).
\end{equation*}
Recalling that the gradient of the log determinant is $\frac{\partial}{\partial A} \log \det A = A^{-1}$, and the gradient $\frac{\partial}{\partial A} x^\top A^{-1} y = -A^{-\top}x y^\top A^{-\top}$ \citep{petersen2008matrix} we can take the gradient of this energy function with respect to the means $\mu$ and covariances $\Sigma$:
\begin{align*}
\frac{\partial \log E(P_i,P_j)}{\partial \mu_i} &= -\frac{\partial \log E(P_i,P_j)}{\partial \mu_j} = -\Delta_{ij}\\
\frac{\partial \log E(P_i,P_j)}{\partial \Sigma_i} &= \frac{\partial \log E(P_i,P_j)}{\partial \Sigma_j} = \frac{1}{2}(\Delta_{ij} \Delta_{ij}^\top - (\Sigma_i + \Sigma_j)^{-1})
\\ & \text{where}~~~~
\Delta_{ij} = (\Sigma_i + \Sigma_j)^{-1}(\mu_i - \mu_j)
\end{align*}
% \begin{align*}
% \frac{\partial E(P_i,P_j)}{\partial \mu_i} &= -\frac{\partial E(P_i,P_j)}{\partial \mu_j} = \Delta_{ij},~~~~
% \frac{\partial E(P_i,P_j)}{\partial \Sigma_i} = \frac{\partial E(P_i,P_j)}{\partial \Sigma_j} = -\frac{1}{2}(\Delta_{ij} \Delta_{ij}^\top - (\Sigma_i + \Sigma_j)^{-1})
% \\ & \text{where}~~~~
% \Delta_{ij} = (\Sigma_i + \Sigma_j)^{-1}(\mu_i - \mu_j)
% \end{align*}
For diagonal and spherical covariances, these matrix inverses are trivial to compute, and even in the full-matrix case can be solved very efficiently for the small dimensionality common in embedding models. If the matrices have a low-rank plus diagonal structure, they can be computed and stored even more efficiently using the matrix inversion lemma.

This log-energy has an intuitive geometric interpretation as a similarity measure. Gaussians are measured as close to one another based on the distance between their means, as measured through the Mahalanobis distance defined by their joint inverse covariance. Recalling that $\log \det A + \text{const.}$ is equivalent to the log-volume of the ellipse spanned by the principle components of $A$, we can interpret this other term of the energy as a regularizer that prevents us from decreasing the distance by only increasing joint variance. This combination pushes the means together while encouraging them to have more concentrated, sharply peaked distributions in order to have high energy.

\subsection{Asymmetric Similarity: KL Divergence}

Training vectors through KL-divergence to encode their context
distributions, or even to incorporate more explicit directional supervision re: entailment from a knowledge base or WordNet,  is also a sensible objective choice. We optimize the
following energy function (which has a similarly tractable closed form solution
for Gaussians):
\begin{align*}
-E(P_i,P_j)&=D_{KL}(\mathcal{N}_j || \mathcal{N}_i) = \int_{x \in \mathbb{R}^n}\mathcal{N}(x;\mu_i,\Sigma_i)\log \frac{\mathcal{N}(x;\mu_j,\Sigma_j)}{\mathcal{N}(x;\mu_i,\Sigma_i)}dx\\
&= \frac{1}{2}(\tr(\Sigma_i^{-1}\Sigma_j) + (\mu_i - \mu_j)^\top \Sigma_i^{-1}(\mu_i - \mu_j) -d - \log \frac{\det(\Sigma_j)}{\det(\Sigma_i)})
\end{align*}
Note the leading negative sign (we define the negative energy), since KL is a distance function and not a similarity. KL divergence is a natural energy function for representing entailment between concepts -- a low KL divergence from $x$ to $y$ indicates that we can encode $y$ easily as $x$, implying that $y$ entails $x$. This can be more intuitively visualized and interpreted as a soft form of inclusion between the level sets of ellipsoids generated by the two Gaussians -- if there is a relatively high expected log-likelihood ratio (negative KL), then most of the mass of $y$ lies inside $x$.

Just as in the previous case, we can compute the gradients for this energy function in closed form:
\begin{align*}
\frac{\partial E(P_i,P_j)}{\partial \mu_i} &= -\frac{\partial E(P_i,P_j)}{\partial \mu_j} = -\Delta'_{ij}\\
\frac{\partial E(P_i,P_j)}{\partial \Sigma_i} &= \frac{1}{2}( \Sigma_i^{-1} \Sigma_j \Sigma_i^{-1} + \Delta'_{ij}\Delta_{ij}^{'\top} - \Sigma_i^{-1} )\\
\frac{\partial E(P_i,P_j)}{\partial \Sigma_j} &= \frac{1}{2}( \Sigma_j^{-1} - \Sigma_i^{-1} )
\\ & \text{where}~~~~
\Delta'_{ij} = \Sigma_i^{-1}(\mu_i - \mu_j)
\end{align*}
% \vspace{0.5cm}
\\using the fact that $\frac{\partial}{\partial A} \tr(X^\top A^{-1} Y) = -(A^{-1} YX^\top A^{-1})^\top$ and $\frac{\partial}{\partial A} \tr (XA) = X^\top$ \citep{petersen2008matrix}.

\subsection{Uncertainty of Inner Products}

Another benefit of embedding objects as probability distributions is that we can look at the distribution of dot products between vectors drawn from two Gaussian representations. This distribution is not itself a one-dimensional Gaussian, but it has a finite mean and variance with a simple structure in the case where the two Gaussians are assumed independent \citep{brown1977means}. For the distribution $P(z=x^\top y)$, we have
\begin{align*}
\mu_z &= \mu_x^\top \mu_y\\
\Sigma_z &=\mu_x^\top \Sigma_x \mu_x + \mu_y^\top \Sigma_y \mu_y + \text{tr}(\Sigma_x \Sigma_y)
\end{align*}
this means we can find e.g. a lower or upper bound on the dot products of random samples from these distributions, that should hold some given percent of the time. Parametrizing this energy by some number of standard deviations $c$, we can also get a range for the dot product as:
\begin{equation*}
\mu_x^\top \mu_y \pm c \sqrt{\mu_x^\top \Sigma_x \mu_x + \mu_y^\top \Sigma_y \mu_y + \text{tr}(\Sigma_x \Sigma_y)}
\end{equation*}
We can choose $c$ in a principled using an (incorrect) Gaussian approximation, or more general concentration bounds such as Chebyshev's inequality.

\subsection{Learning}

To learn our model, we need to pick an energy function (EL or KL), a loss function (max-margin), and a set of positive and negative training pairs. As the landscape is highly nonconvex, it is also helpful to add some regularization.

We regularize the means and covariances differently, since they are different types of geometric objects. The means should not be allowed to grow too large, so we can add a simple hard constraint to the $\ell_2$ norm:
\begin{equation*}
\|\mu_i\|_2 \le C, ~~\forall i
\end{equation*}
However, the covariance matrices need to be kept positive definite as well as reasonably sized. This is achieved by adding a hard constraint that the eigenvalues $\lambda_i$ lie within the hypercube $[m, M]^d$ for constants $m$ and $M$.
\begin{equation*}
mI \prec \Sigma_i \prec MI, ~~\forall i
\end{equation*}
For diagonal covariances, this simply involves either applying the min or max function to each element of the diagonal to keep it within the hypercube, $\Sigma_{ii} \leftarrow \max(m, \min(M,\Sigma_{ii}))$.

Controlling the bottom eigenvalues of the covariance is especially important when training with expected likelihood, since the energy function includes a $\log \det$ term that can give very high scores to small covariances, dominating the rest of the energy.

We optimize the parameters using AdaGrad \citep{duchi2011adaptive} and stochastic gradients in small minibatches containing 20 sentences worth of tokens and contexts.

\section{Evaluation}
\label{evaluation}

We evaluate the representation learning algorithms on several qualitative and quantitative tasks, including modeling asymmetric and linguistic relationships, uncertainty, and word similarity. All Gaussian experiments are conducted with 50-dimensional vectors, with diagonal variances except where noted otherwise. Unsupervised embeddings are learned on the concatenated ukWaC and WaCkypedia corpora \citep{baroni2009wacky}, consisting of about 3 billion tokens. This matches the experimental setup used by \cite{Baroni:2012:EAW:2380816.2380822}, aside from leaving out the small British National Corpus, which is not publicly available and contains only 100 million tokens. All word types that appear less than 100 times in the training set are dropped, leaving a vocabulary of approximately 280 thousand word types.

When training word2vec Skip-Gram embeddings for baselines, we follow the above training setup (50 dimensional embeddings), using our own implementation of word2vec to change as little as possible between the two models, only the loss function. We train both models with one pass over the data, using separate embeddings for the input and output contexts, 1 negative sample per positive example, and the same subsampling procedure as in the word2vec paper \citep{mikolov2013distributed}. The only other difference between the two training regimes is that we use a smaller $\ell_2$ regularization constraint when using the word2vec loss function, which improves performance vs. the diagonal Gaussian model which does better with ``spikier'' mean embeddings with larger norms (see the comment in Section \ref{word-sim-section}). The original word2vec implementation uses no $\ell_2$ constraint, but we saw better performance when including it in our training setup.

\subsection{Specificity and Uncertainty of Embeddings}

In Figure \ref{specificity-table}, we examine some of the 100 nearest neighbors of several query words as we sort from largest to smallest variance, as measured by determinant of the covariance matrix, using diagonal Gaussian embeddings. Note that more specific words, such as \emph{joviality} and \emph{electroclash} have smaller variance, while polysemous words or those denoting broader concepts have larger variances, such as \emph{mix}, \emph{mind}, and \emph{graph}. This is not merely an artifact of higher frequency words getting more variance -- when sorting by those words whose rank by frequency and rank by variance are most dissimilar, we see that genres with names like \emph{chillout}, \emph{avant}, and \emph{shoegaze} overindex their variance compared to how frequent they are, since they appear in different contexts. Similarly, common emotion words like \emph{sadness} and \emph{sincerity} have less variance than their frequency would predict, since they have fairly fixed meanings. Another emotion word, \emph{coldness}, is an uncommon word with a large variance due to its polysemy.

\begin{figure}[t!]
    \centering
\begin{tabular}[t]{|ll|}
\hline
\bf Query Word & \bf Nearby Words, Descending Variance \\\hline\hline
rock & mix sound blue folk jazz rap avant hardcore chillout shoegaze powerpop \\ & electroclash\\
food & drink meal meat diet spice juice bacon soya gluten stevia\\
feeling & sense mind mood perception compassion sadness coldness sincerity\\ & perplexity diffidence joviality\\
algebra & theory graph equivalence finite predicate congruence topology \\ & quaternion symplectic homomorphism\\
\hline
\end{tabular}
\caption{\label{specificity-table}Elements of the top 100 nearest neighbor sets for chosen query words, sorted by descending variance (as measured by determinant of covariance matrix). Note that less specific and more ambiguous words have greater variance.}
\end{figure}

\subsection{Entailment}
\label{entailment-experiments}

% As can be seen qualitatively in Figures \ref{fig-ellipses} and \ref{fig-ratel}, our embeddings can learn some 
As can be seen qualitatively in Figure \ref{fig-ellipses}, our embeddings can learn some 
forms of unsupervised entailment directly from the source data. We evaluate quantitatively on the Entailment dataset of \cite{Baroni:2012:EAW:2380816.2380822}. Our setup is essentially the same as theirs but uses slightly less data, as mentioned in the beginning of this section. We evaluate with Average Precision and best F1 score. We include the best F1 score (by picking the optimal threshold at test) because this is used by \cite{Baroni:2012:EAW:2380816.2380822}, but we believe AP is better to demonstrate the correlation of various asymmetric and symmetric measures with the entailment data.

\begin{figure}[t!]
    \centering
\begin{tabular}[t]{|lllll|}
\hline
\bf Model & \bf Test & \bf Similarity & \bf Best F1 & \bf AP \\\hline\hline
\cite{Baroni:2012:EAW:2380816.2380822} & E & balAPinc & \bf 75.1 & --\\\hline
Empirical (D) & E & KL & 70.05 & .68\\
Empirical (D) & E & Cos & 76.24 & .71\\
Empirical (S) & E & KL & 71.18 & .69\\
Empirical (S) & E & Cos & 76.24 & .71\\
Learned (D) & E & KL & 79.01 & \bf .80\\
Learned (D) & E & Cos & 76.99 & .73\\
Learned (S) & E & KL & \bf 79.34 & .78\\
Learned (S) & E & Cos & 77.36 & .73\\\hline
\end{tabular}\\
\caption{\label{entailment-table}Entailment: We compare empirical and learned variances, both diagonal (D) and spherical (S). E is the dataset of \cite{Baroni:2012:EAW:2380816.2380822}.
Measures of similarity are symmetric (cosine between means) and asymmetric (KL) divergence for Gaussians. balAPinc is an asymmetric similarity measure specific to sparse, distributional count-based representations.}
\end{figure}

In Figure \ref{entailment-table}, we compare variances learned jointly during embedding training by using the expected likelihood objective, with empirical variances gathered from contexts on pre-trained word2vec-style embeddings. We compare both diagonal (D) and spherical (S) variances, using both cosine similarity between means, and KL divergence. Baseline asymmetric measurements, such as the difference between the sizes of the two embeddings, did worse than the cosine. We see that KL divergence between the entailed and entailing word does not give good performance for the empirical variances, but beats the count-based balAPinc measure when used with learned variances. 

For the baseline empirical model to achieve reasonable performance when using KL divergence, we regularized the covariance matrices, as the unregularized matrices had very small entries. We regularized the empirical covariance by adding a small ridge $\delta$ to the diagonal, which was tuned to maximize performance, to give the largest possible advantage to the baseline model. Interestingly, the empirical variances do worse with KL than the symmetric cosine similarity when predicting entailment. This appears to be because the empirically learned variances are so small that the choice is between either leaving them small, making it very difficult to have one Gaussian located ``inside'' another Gaussian, or regularizing so much that their discriminative power is washed out. Additionally, when examining the empirical variances, we noted that common words like ``such,'' which receive very large variances in our learned model, have much smaller empirical variances relative to rarer words. A possible explanation is that the contrastive objective forces variances of commonly sampled words to spread out to avoid loss, while the empirical variance sees only ``positive examples'' and has no penalty for being close to many contexts at once.

While these results indicate that we can do as well or better at unsupervised entailment than previous distributional semantic measures, we would like to move beyond purely unsupervised learning. Although certain forms of entailment can be learned in an unsupervised manner from distributional data, many entailing relationships are not present in the training text in the form of lexical substitutions that reflect the \emph{is-a} relationship. For example, one might see phrases such as ``look at that bird,'' ``look at that eagle,'' ``look at that dog,'' but rarely ``look at that mammal.'' One appealing aspect of our models versus count-based ones is that they can be directly discriminatively trained to embed hierarchies.

\begin{figure}[t!]
    \centering

\begin{subfigure}[t]{0.5\textwidth}
\includegraphics[width=\textwidth]{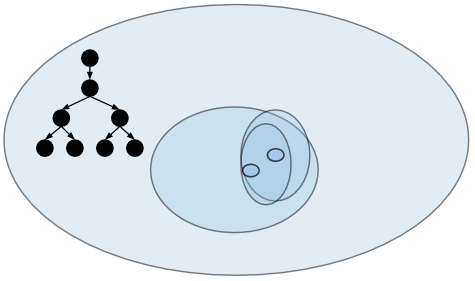}
\end{subfigure}~
\begin{subfigure}[t]{0.5\textwidth}
\includegraphics[scale=0.43]{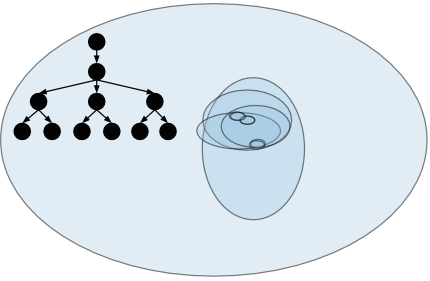}

\end{subfigure}%

\caption{\label{fig-synthetic-hierarchy}Synthetic experiments on embedding two simple hierarchies in two dimensions directly using KL divergence. The embedding model captures all of the hierarchical relationships present in the tree. Sibling leaves are pushed into overlapping areas by the objective function.}

\end{figure}

\subsection{Directly Learning Asymmetric Relationships}

In Figure \ref{fig-synthetic-hierarchy}, we see the results of directly embedding simple tree hierarchies as Gaussians. We embed nodes as Gaussians with diagonal variances in two-dimensional space using gradient descent on the KL divergence between parents and children. We create a Gaussian for each node in the tree, and randomly initialize means. Negative contexts come from randomly sampled nodes that are neither ancestors nor descendents, while positive contexts come from ancestors or descendents using the appropriate directional KL divergence. Unlike our experiments with symmetric energy, we must use the same set of embeddings for nodes and contexts, or else the objective function will push the variances to be unboundedly large. Our training process captures the hierarchical relationships, although leaf-level siblings are not differentiated from each other by this objective function. This is because out of all the negative examples that a leaf node can receive, only one will push it away from its sibling node.

\subsection{Word Similarity Benchmarks}
\label{word-sim-section}

We evaluate the embeddings on seven different standard word similarity benchmarks \citep{Rubenstein:1965:CCS:365628.365657,szumlanski2013new,hill2014simlex,miller1991contextual,bruni2014multimodal,Yang06verbsimilarity,finkelstein2001placing}. A comparison to all of the state of the art word-embedding numbers for different dimensionalities as in \citep{P14-1023} is out of the scope of this evaluation. However, we note that the overall performance of our 50-dimensional embeddings matches or beats reported numbers on these datasets for the 80-dimensional Skip-Gram vectors at \emph{wordvectors.org} \citep{faruqui-2014:SystemDemo}, as well as our own Skip-Gram implementation. Note that the numbers are not directly comparable since we use a much older version of Wikipedia (circa 2009) in our WaCkypedia dataset, but this should not give us an edge.

While it is good to sanity-check that our embedding algorithms can achieve standard measures of distributional quality, these experiments also let us compare the different types of variances (spherical and diagonal). We also compare against Skip-Gram embeddings with 100 latent dimensions, since our diagonal variances have 50 extra parameters.

We see that the embeddings with spherical covariances have an overall slight edge over the embeddings with diagonal covariances in this case, in a reversal from the entailment experiments. This could be due to the diagonal variance matrices making the embeddings more axis-aligned, making it harder to learn all the similarities and reducing model capacity. To test this theory, we plotted the absolute values of components of spherical and diagonal variance mean vectors on a q-q plot and noted a significant off-diagonal shift, indicating that diagonal variance embedding mean vectors have ``spikier'' distributions of components, indicating more axis-alignment.

We also see that the distributions with diagonal variances benefit more from including the variance in the comparison (d) than the spherical variances. Generally, the data sets in which the cosine between distributions (d) outperforms cosine between means (m) are similar for both spherical and diagonal covariances. Using the cosine between distributions never helped when using empirical variances, so we do not include those numbers.

\begin{figure}[t!]
    \centering
\begin{tabular}[t]{|ll|l|ll|ll|}
\hline
\bf Dataset & \bf SG (50d) & \bf SG (100d) & \bf LG/50/m/S & \bf LG/50/d/S & \bf LG/50/m/D & \bf LG/50/d/D\\\hline\hline
SimLex & 29.39 & 31.13 & \bf 32.23 & 29.84 & \bf 31.25 & 30.50\\
WordSim & 59.89 & 59.33 & \bf 65.49 & 62.03 & \bf 62.12 & 61.00\\
WordSim-S & 69.86 & 70.19 & \bf 76.15 & 73.92 & \bf 74.64 & 72.79\\
WordSim-R & 53.03 & 54.64 & \bf 58.96 & 54.37 & \bf 54.44 & 53.36\\
MEN & 70.27 & 70.70 & \bf 71.31 & 69.65 & \bf 71.30 & 70.18\\
MC & 63.96 & 66.76 & \bf 70.41 & 69.17 & 67.01 & \bf 68.50\\
RG & 70.01 & 69.38 & 71.00 & \bf74.76 & 70.41 & \bf 77.00\\
YP & 39.34 & 35.76 & 41.50 & \bf 42.55 & 36.05 & \bf 39.30\\
Rel-122 & 49.14 & 51.26 & \bf 53.74 & 51.09 & 52.28 & \bf 53.54\\\hline
\end{tabular}\\
\caption{Similarity: We evaluate our learned Gaussian embeddings (LG) with spherical (S) and diagonal (D) variances, on several word similarity benchmarks, compared against standard Skip-Gram (SG) embeddings on the trained on the same dataset. We evaluate Gaussian embeddings with both cosine between means (m), and cosine between the distributions themselves (d) as defined by the expected likelihood inner product.\label{similarity-table}}
\end{figure}

\section{Conclusion and Future Work}

In this work we introduced a method to embed word types into the space of Gaussian distributions, and
learn the embeddings directly in that space. This allows us to represent words not as low-dimensional vectors, but as densities over a latent space, directly representing notions of uncertainty and enabling a richer geometry in the embedded space. We demonstrated the effectiveness of these embeddings on a linguistic task requiring asymmetric comparisons, as well as standard word similarity benchmarks, learning of synthetic hierarchies, and several qualitative examinations.

In future work, we hope to move beyond spherical or diagonal covariances and into combinations of low rank and diagonal matrices. Efficient updates and scalable learning is still possible due to the Sherman-Woodbury-Morrison formula. Additionally, going beyond diagonal covariances will enable us to keep our semantics from being axis-aligned, which will increase model capacity and expressivity. We also hope to move past stochastic gradient descent and warm starting and be able to learn the Gaussian representations robustly in one pass from scratch by using e.g. proximal or block coordinate descent methods. Improved optimization strategies will also be helpful on the highly nonconvex problem of training supervised hierarchies with KL divergence.

Representing words and concepts as different types of distributions (including other elliptic distributions such as the Student's t) is an exciting direction -- Gaussians concentrate their density on a thin spherical ellipsoidal shell, which can lead to counterintuitive behavior in high dimensions. Multimodal distributions represent another clear avenue for future work. Combining ideas from kernel methods and manifold learning with deep learning and linguistic representation learning is an exciting frontier.

In other domains, we want to extend the use of potential function representations to other tasks requiring embeddings, such as relational learning with the universal schema \citep{riedel2013relation}. We hope to leverage the asymmetric measures, probabilistic interpretation, and flexible training criteria of our model to tackle tasks involving similarity-in-context, comparison of sentences and paragraphs, and more general common sense reasoning. 

\section{Acknowledgements}

This work was supported in part by the Center for Intelligent Information Retrieval, in part by IARPA via DoI/NBC contract \#D11PC20152, and in part by NSF grant \#CNS-0958392 The U.S. Government is authorized to reproduce and distribute reprint for Governmental purposes notwithstanding any copyright annotation thereon. Any opinions, findings and conclusions or recommendations expressed in this material are those of the authors and do not necessarily reflect those of the sponsor.

\bibliography{paper}
\bibliographystyle{iclr2015}

\end{document}